\documentclass[10pt,twocolumn,letterpaper]{article}

\usepackage{cvpr}
\usepackage{times}
\usepackage{epsfig}
\usepackage{graphicx}
\usepackage{amsmath}
\usepackage{amssymb}

\usepackage[pagebackref=true,breaklinks=true,letterpaper=true,colorlinks,bookmarks=false]{hyperref}

\usepackage{subfigure}
\usepackage{booktabs}
\usepackage{pifont}% http://ctan.org/pkg/pifont
\newcommand{\cmark}{\ding{51}}%
\newcommand{\xmark}{\ding{55}}%
\usepackage{multirow}
\usepackage{threeparttable}

\cvprfinalcopy % *** Uncomment this line for the final submission

\begin{document}

%%%%%%%%% TITLE
\title{Learning Dynamic Routing for Semantic Segmentation}

\author{Yanwei Li$^{1,2}$, Lin Song$^{3}$, Yukang Chen$^{1,2}$, Zeming Li$^{4}$, Xiangyu Zhang$^{4}$, \\
	 Xingang Wang$^{1}$, Jian Sun$^{4}$  \\
$^{1}$Institute of Automation, Chinese Academy of Sciences\\
$^{2}$University of Chinese Academy of Sciences\\
$^{3}$Xi'an Jiaotong University
$^{4}$Megvii Technology\\
%{\tt\small \{liyanwei2017,chenyukang2017,xingang.wang\}@ia.ac.cn}\\
%{\tt\small stevengrove@stu.xjtu.edu.cn,\{lizeming,zhangxiangyu,sunjian\}@megvii.com}
}

\maketitle
%\thispagestyle{empty}

%%%%%%%%% ABSTRACT
\begin{abstract}
Recently, numerous handcrafted and searched networks have been applied for semantic segmentation. However, previous works intend to handle inputs with various scales in pre-defined static architectures, such as FCN, U-Net, and DeepLab series. This paper studies a conceptually new method to alleviate the scale variance in semantic representation, named dynamic routing. The proposed framework generates data-dependent routes, adapting to the scale distribution of each image. To this end, a differentiable gating function, called soft conditional gate, is proposed to select scale transform paths on the fly. In addition, the computational cost can be further reduced in an end-to-end manner by giving budget constraints to the gating function. We further relax the network level routing space to support multi-path propagations and skip-connections in each forward, bringing substantial network capacity. To demonstrate the superiority of the dynamic property, we compare with several static architectures, which can be modeled as special cases in the routing space. Extensive experiments are conducted on Cityscapes and PASCAL VOC 2012 to illustrate the effectiveness of the dynamic framework. Code is available at \href{https://github.com/yanwei-li/DynamicRouting}{https://github.com/yanwei-li/DynamicRouting}.\footnote{Work was done in Megvii Research. Email: \href{mailto:liyanwei2017@ia.ac.cn}{liyanwei2017@ia.ac.cn}}
\end{abstract}

%%%%%%%%% BODY TEXT
\section{Introduction}\label{sec:intro}

%------------------------------------------------------------------------
\begin{figure}[t!]
  \centering
    \subfigure[Network architecture of {\em large-scale} input]{
    \includegraphics[width=0.99\linewidth]{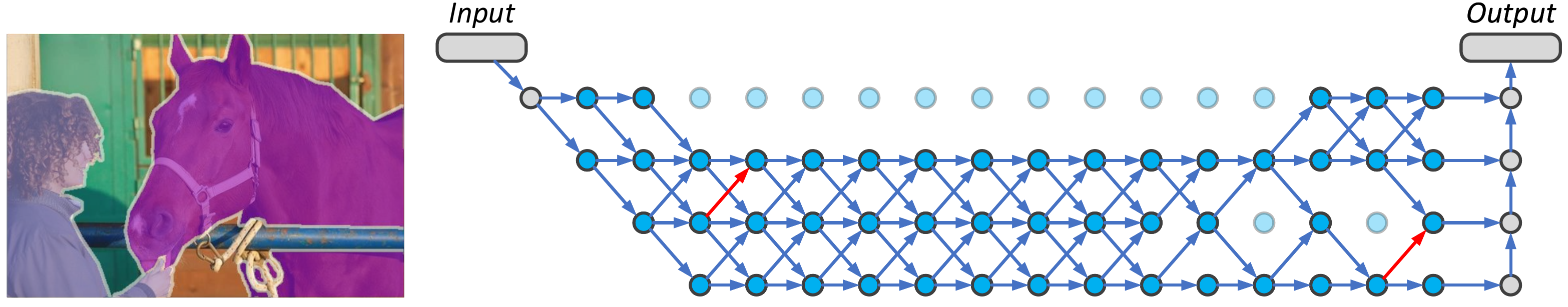}
    \label{fig:arch_intro_large}
    }
    \subfigure[Network architecture of {\em small-scale} input]{
    \includegraphics[width=0.99\linewidth]{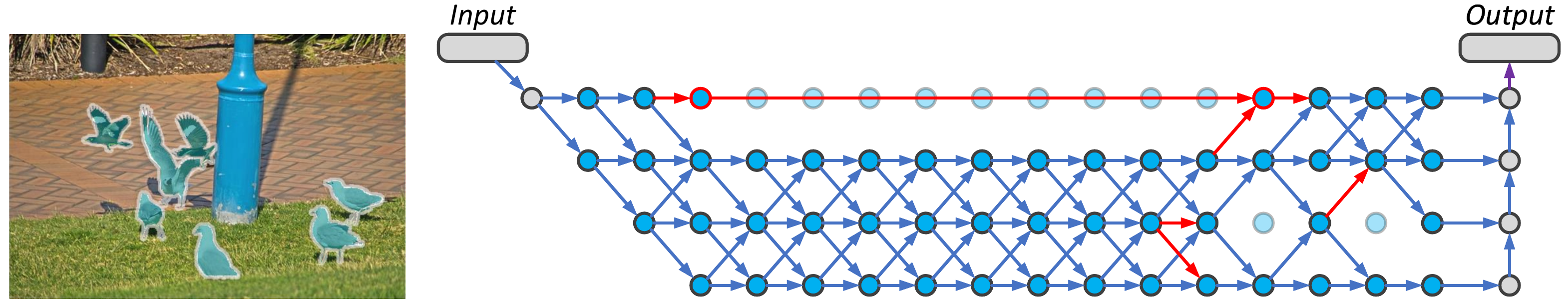}
    \label{fig:arch_intro_small}
    }
    \subfigure[Network architecture of {\em mixed-scale} input]{
    \includegraphics[width=0.99\linewidth]{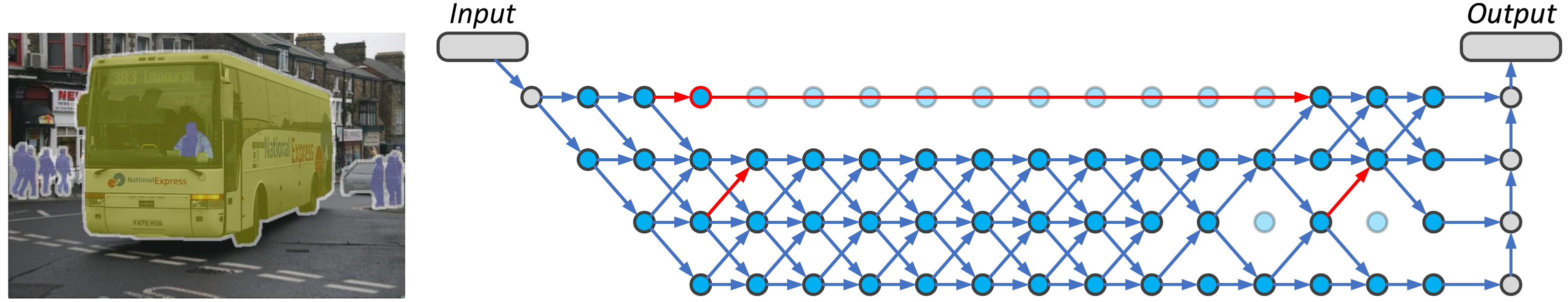}
    \label{fig:arch_intro_mix}
    }
\caption{Given inputs with different scale distributions, the proposed dynamic routing will choose corresponding forward paths. For example, the architecture of {\em large-scale} instances~\ref{fig:arch_intro_large} could ignore low-level features. The {\em small-scale} objects~\ref{fig:arch_intro_small} may depend on low-level details as well as higher resolution. And the {\em mixed-scale} things~\ref{fig:arch_intro_mix} would enjoy both connection patterns. \textcolor{red}{Red} lines in diagrams denote the difference among them. }
% {\bf All these architectures are generated by the proposed approach directly}.}
\label{fig:arch_intro}
\end{figure}
%-------------------------------------------------------------------------

Semantic segmentation, which aims at assigning each pixel with semantic categories, is one of the most fundamental yet challenging tasks in the computer vision field. One of the problems in semantic segmentation comes from the huge scale variance among inputs, {\em e.g.,} the tiny object instances and the picture-filled background stuff. Moreover, the large distribution variance brings difficulties to feature representation as well as relationship modeling. Traditional methods try to solve this problem by well-designed network architectures. For instance, multi-resolution fusion~\cite{long2015fully, ronneberger2015u, badrinarayanan2017segnet, sun2019high, kirillov2019panoptic} is adopted for detail-oriented feature maps, and long-range dependencies are captured for global context modeling~\cite{zhao2017pyramid, wang2018non, zhao2018psanet, chen2018encoder, song2019learnable}. With the development of Neural Architecture Search~(NAS), there are several works delving in searching effective architectures for semantic segmentation automatically~\cite{chen2018searching, liu2019auto, nekrasov2019fast}. 

However, both classic human-designed and NAS-based networks intend to represent all of the instances in a single network architecture, which lacks the adaptability to diverse scale distributions in the real-world environment. An example is presented in Fig.~\ref{fig:arch_intro}, where the scale of instances varies a lot. To this end, a more {\em customizable} network is needed to accommodate the scale variance of each image. 

In this paper, we propose a framework that is conceptually novel for semantic segmentation, called {\em dynamic routing}. In particular, the dynamic routing generates {\em data-dependent} forward paths during inference, which means the specific network architecture varies with inputs. With this method, instances (or backgrounds) with different scales could be allocated to corresponding resolution stages for customized feature transformation. As illustrated in Fig.~\ref{fig:arch_intro}, the input images with diverse scale distributions will choose different routes for feature transformation. There are some researches on dynamic networks for efficient object recognition via dropping blocks~\cite{wu2018blockdrop, huang2017multi, teja2018hydranets, wang2018skipnet} or pruning channels ~\cite{you2019gate, lin2017runtime}. Different from them, this work focuses on {\em semantic representation} and intends to {\em alleviate scale variance as well as improve network efficiency}.

The routing space in traditional dynamic approaches for image classification~\cite{huang2017multi, teja2018hydranets, wang2018skipnet} are usually limited to a resolution declining pipeline, which would not suffice for semantic segmentation. We draw inspiration from the search space of Auto-DeepLab~\cite{liu2019auto} and develop a new routing space for better capacity, which contains several independent {\em cells}. Specifically, different from Auto-DeepLab, {\em multi-path propagations} and {\em skip-connections}, which are proved to be quiet essential in semantic segmentation~\cite{ronneberger2015u, chen2018encoder}, are enabled in each forward during inference. Therefore, several classic network architectures can be included as special cases for comparisons (Fig.~\ref{fig:arch_in_space}). In terms of the dynamic routing, a {\em data-dependent} routing gate, called {\em soft conditional gate}, is designed to select each path according to the input image. With the proposed routing gate, each basic cell, as well as the resolution transformation path, can be taken into consideration individually. Moreover, the proposed routing gate can be formulated into a {\em differentiable} module for end-to-end optimization. Consequently, given limited computational budgets ({\em e.g.,} FLOPs), cells with little contribution will be dropped on the fly. 

The overall approach, named {\em dynamic routing}, can be easily instantiated for semantic segmentation. To elaborate on its superiority over the fixed architectures in both performance and efficiency, we give extensive ablation studies and detailed analyses in Sec.~\ref{sec:exper_dynamic_routing}. Experimental results are further reported on two well-known datasets, namely Cityscapes~\cite{cordts2016cityscapes} and PASCAL VOC 2012~\cite{everingham2010pascal}. With the simple scale transformation modules, the proposed dynamic routing achieves comparable results with the state-of-the-art methods but consumes much fewer resources.

%-------------------------------------------------------------------------
\section{Related Works}\label{sec:related_works}
Traditional semantic segmentation researches mainly focused on designing subtle network architectures by human experiences~\cite{long2015fully, ronneberger2015u, badrinarayanan2017segnet, zhao2017pyramid, chen2018encoder}. With the development of NAS, there are several methods attempting to search for a {\em static} network automatically~\cite{chen2018searching, liu2019auto, nekrasov2019fast}. 
%However, due to the variance in scale distribution, static networks could not settle all the circumstances. 
Unlike previous works, the {\em dynamic routing} is proposed to select the most suitable scale transform according to the input, which has seldom been explored. Herein, we first retrospect hand-designed architectures for semantic segmentation. Then we give an introduction to NAS-based approaches. Finally, previous developments of dynamic networks are reviewed.
%-------------------------------------------------------------------------
\subsection{Handcrafted Architectures}
Handcrafted architectures have been well studied in recent years. There are several researches delving in network design for semantic segmentation, {\em e.g.,} FCN~\cite{long2015fully}, U-Net~\cite{ronneberger2015u}, Conv-Deconv~\cite{noh2015learning}, SegNet~\cite{badrinarayanan2017segnet}. Based on the well-designed FCN~\cite{long2015fully} and U-shape architecture~\cite{ronneberger2015u}, numerous works have been proposed to model global context by capturing larger receptive field~\cite{zhao2017pyramid, chen2017deeplab, chen2017rethinking, chen2018encoder, yu2018learning} or establishing pixel-wise relationships~\cite{zhao2018psanet, huang2018ccnet, fu2019dual, song2019learnable}. Due to the high resource consumption of dense prediction, some light-weighted architectures have been proposed for the sake of efficiency, including ICNet~\cite{zhao2018icnet} and BiSeNet~\cite{yu2018bisenet}. Overall, handcrafted architectures aim at utilizing multi-scale features from different stages in a static network, rather than adapting to input dynamically.
%-------------------------------------------------------------------------
\subsection{NAS-based Approaches}
Recently, Neural Architecture Search (NAS) has been widely used for automatic network architecture design~\cite{zoph2016neural, pham2018efficient, liu2018darts, cai2018proxylessnas, guo2019single, chen2019progressive}. When it comes to the specific domain, there are several approaches trying to search for effective architectures that are more suitable for semantic segmentation. Specifically, Chen {\em et.al.}~\cite{chen2018searching} searches for multi-scale module to replace ASPP~\cite{chen2017rethinking} block. Furthermore, Nekrasov {\em et.al.}~\cite{nekrasov2019fast} studies the routing type of auxiliary cells in the decoder using the NAS-based method. More recently, Auto-DeepLab~\cite{liu2019auto} is proposed to search for a single route from the dense-connected search space. Different from the NAS-based approaches, which search for a single architecture and then retrain it, the proposed dynamic routing generates forward paths on the fly {\em without searching}.
%-------------------------------------------------------------------------
\begin{figure*}[th!]
  \centering
  \includegraphics[width=\linewidth]{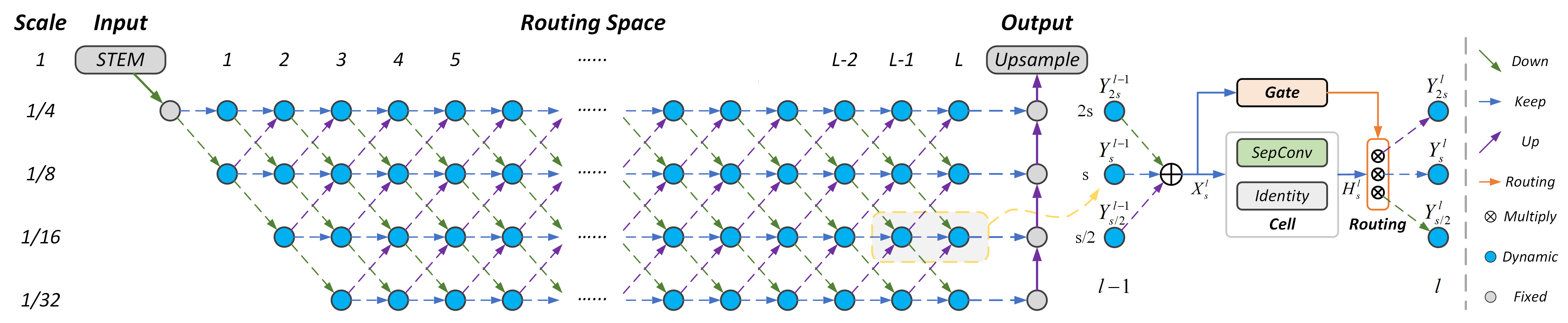}\\
  \caption{The proposed {\em dynamic routing} framework for semantic segmentation. {\em Left:} The routing space with layer $L$ and max down-sampling rate 32. The beginning {\em STEM} and the final {\em Upsample} block are fixed for stability. Dashed lines denote alternative paths for dynamic routing. {\em Right:}  Dynamic routing process at the cell level. Given the summed input from the former layer, we first generate activating weights using the {\em Soft Conditional Gate}. Paths with corresponding weights above zero are marked as activated, which would be selected for feature transformation. More details about the network are elaborated in Sec.~\ref{sec:network_arch}. Best viewed in color.}
  \label{fig:arch}
\end{figure*}
%-------------------------------------------------------------------------
\subsection{Dynamic Networks}
Dynamic networks, adjusting the network architecture to the corresponding input, have been recently studied in the computer vision domain. Traditional methods mainly focus on image classification by dropping blocks~\cite{wu2018blockdrop, huang2017multi, teja2018hydranets, wang2018skipnet} or pruning channels~\cite{you2019gate, lin2017runtime} for efficient inference. For example, an early-existing strategy is adopted in MSDNet~\cite{huang2017multi} for resource-efficient object recognition, which classifies easier inputs and gives output in earlier stages. And SkipNet~\cite{wang2018skipnet} attempts to skip convolutional blocks using an RL-based gating network. However, dynamic routing has seldom been explored for scale transformation, especially in semantic segmentation. To utilize the dynamic property, an end-to-end dynamic routing framework is proposed in this paper to alleviate the scale variance among inputs.
%-------------------------------------------------------------------------
\begin{figure}[t!]
  \centering
    \subfigure[Network architecture modeled from FCN-32s~\cite{long2015fully}]{
    \includegraphics[width=0.98\linewidth]{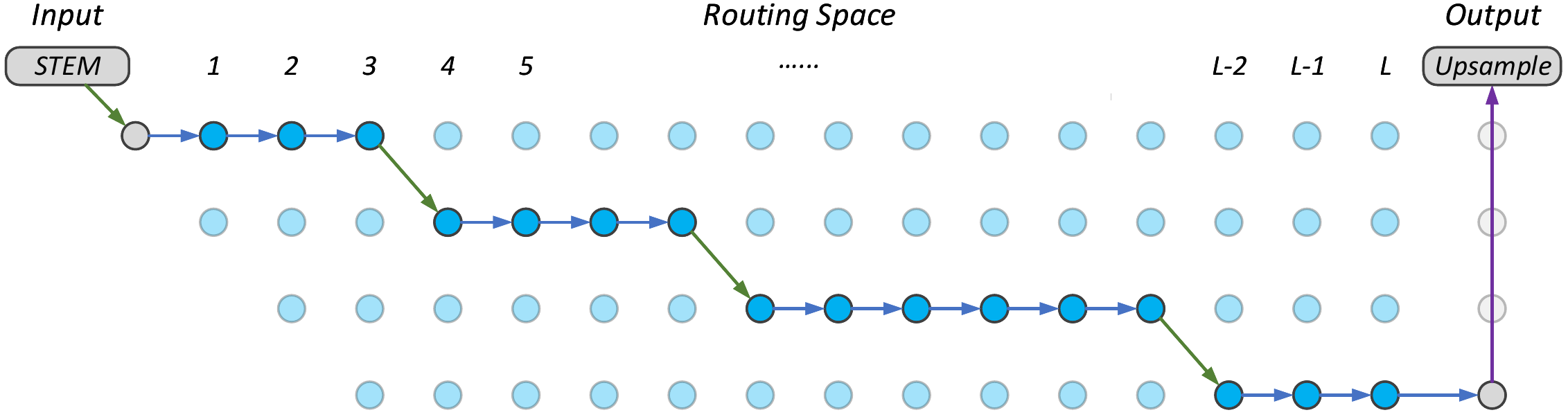}
    \label{fig:archs_in_space_fcn32s}
    }
    \subfigure[Network architecture modeled from U-Net~\cite{ronneberger2015u}]{
    \includegraphics[width=0.98\linewidth]{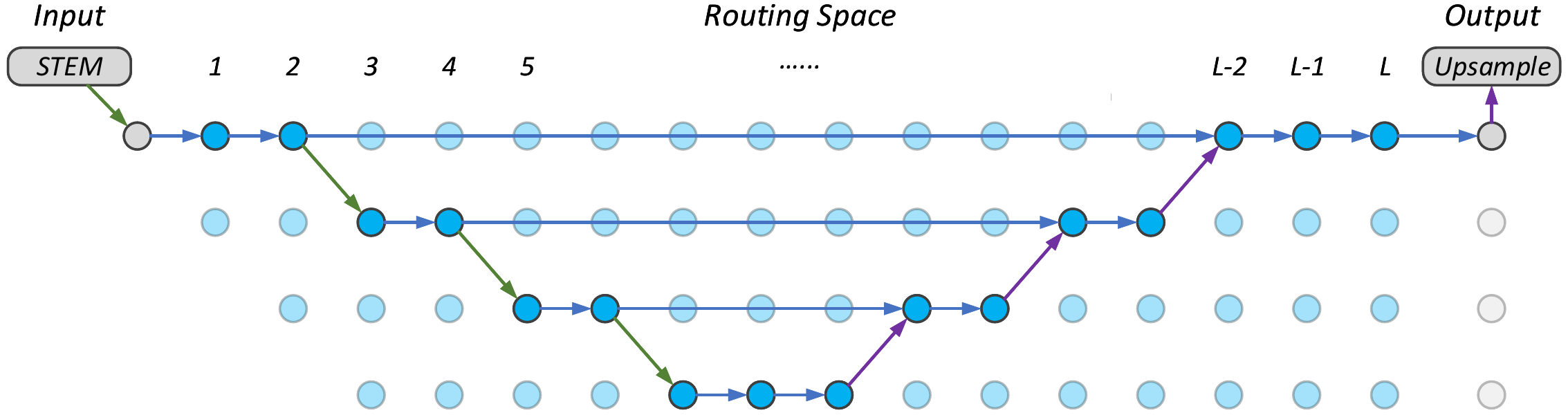}
    \label{fig:archs_in_space_unet}
    }
    \subfigure[Network architecture modeled from DeepLabV3~\cite{chen2017rethinking}]{
    \includegraphics[width=0.98\linewidth]{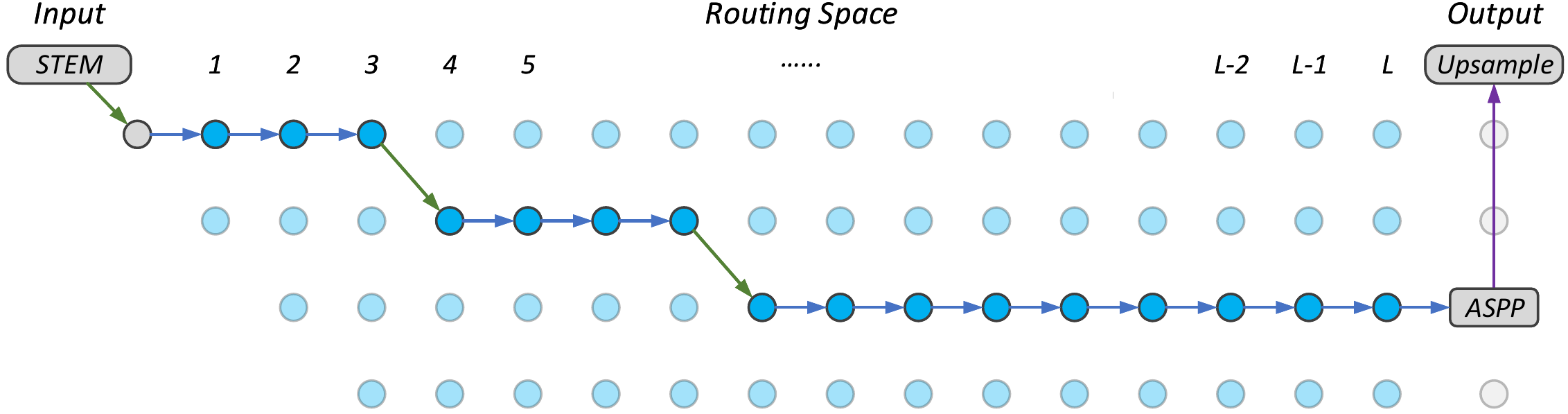}
    \label{fig:archs_in_space_deeplabv3}
    }
    \subfigure[Network architecture modeled from HRNetV2~\cite{sun2019high}]{
    \includegraphics[width=0.98\linewidth]{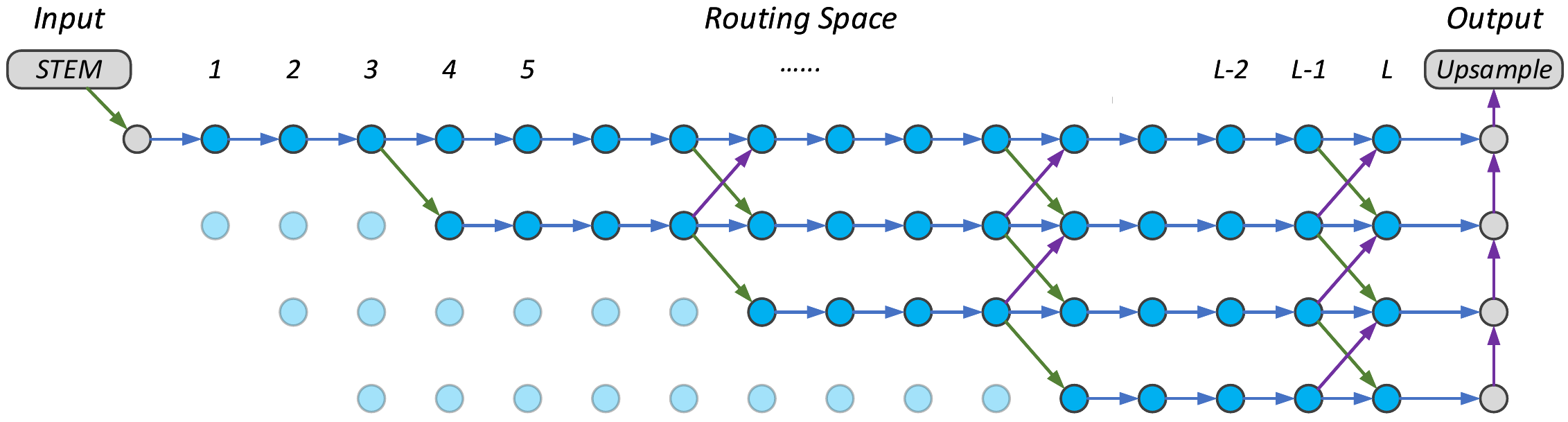}
    \label{fig:archs_in_space_hrnetv2}
    }
    \subfigure[Network architecture modeled from Auto-DeepLab~\cite{liu2019auto}]{
    \includegraphics[width=0.98\linewidth]{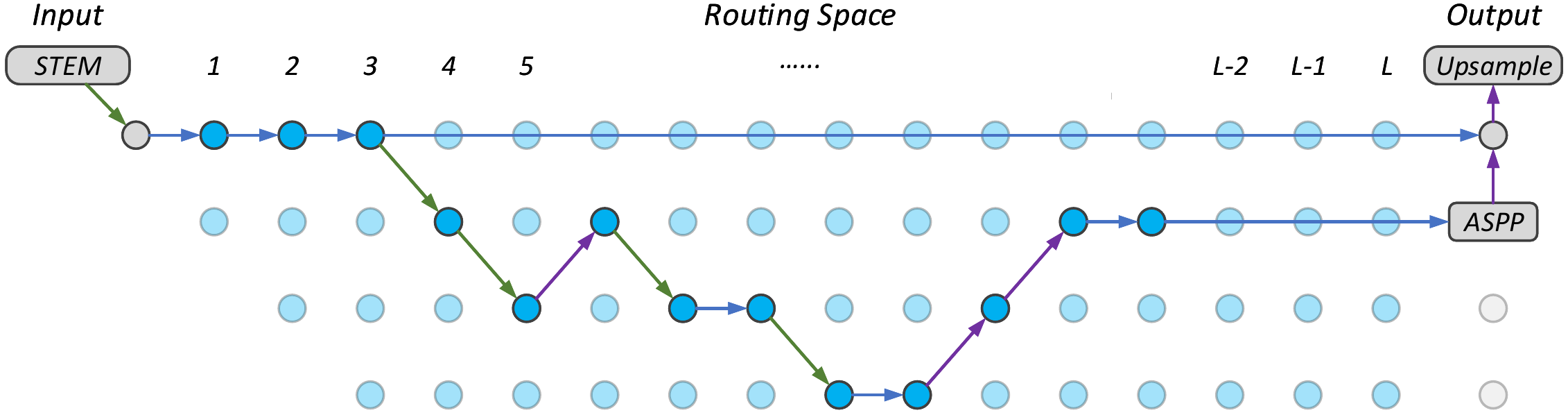}
    \label{fig:archs_in_space_autodeeplab}
    }
\caption{Sampled architectures from previous works. With the designed routing space, several classic architectures can be formulated in similar forms, {\em e.g.,} FCN-32s~\ref{fig:archs_in_space_fcn32s}, U-Net~\ref{fig:archs_in_space_unet}, DeepLabV3~\ref{fig:archs_in_space_deeplabv3}, HRNetV2~\ref{fig:archs_in_space_hrnetv2}, and Auto-DeepLab~\ref{fig:archs_in_space_autodeeplab}.}
\label{fig:arch_in_space}
\end{figure}
%-------------------------------------------------------------------------
\section{Learning Dynamic Routing}\label{sec:dyncamic_routing}
%-------------------------------------------------------------------------
Compared with static network architectures, dynamic routing has the superiority in network capacity and higher performance with the budgeted resource consumption. In this section, we first introduce the designed routing space. And then, the dynamic routing framework and the constraint mechanism are elaborated. The architecture details will be given at the end of this section.
%-------------------------------------------------------------------------
\subsection{Routing Space}\label{sec:routing_space}
To release the potential of dynamic routing, we provide fully-connected paths between adjacent layers with some prior constraints, {\em e.g.,} the up-sampling or down-sampling stride between cells, as illustrated in Fig.~\ref{fig:arch}. Specifically, following the common practices in the network design, the beginning of the network is a fixed 3-layer `STEM' block, which reduces the resolution to 1/4 scale. After that, a space with $L$ layers is designed for dynamic routing, called {\em routing space}. In the routing space, the scaling factor between adjacent cells is restricted to 2, which is widely adopted in ResNet-based methods. Thus, the minimum scale is set to 1/32. With these constraints, the number of candidates in each layer is up to 4. And there are 3 paths for scale transformation in each candidate, namely up-sampling, keeping resolution, and down-sampling. Inside each candidate, basic {\em cell} is designed for feature aggregation, while a fundamental {\em gate} is proposed for path selection, as presented in Fig.~\ref{fig:arch}. The layer-by-layer up-sampling module is fixed at the end of the network to generate predictions. More details about the dynamic routing process are explained in Sec.~\ref{sec:routing_process}.

Different from Auto-DeepLab~\cite{liu2019auto}, where only one specific path in each node is selected in the inference stage, we further relax the routing space to support {\em multi-path} routes and {\em skip-connections} in each candidate. With the more generic space, a lot of popular architectures can be formulated as special cases, as visualized in Fig.~\ref{fig:arch_in_space}. Further quantitative comparisons are given in Sec.~\ref{sec:exper_dynamic_routing} to demonstrate the superiority of the dynamic routing.
%-------------------------------------------------------------------------
\subsection{Routing Process}\label{sec:routing_process}
Given the routing space with several individual nodes, we adopt a basic {\em cell} and a corresponding {\em gate} inside each node to aggregate multi-scale features and choose routing paths, respectively. This process is briefly illustrated in Fig.~\ref{fig:arch}. To be more specific, we first aggregate three inputs with different spatial sizes (namely, $s/2$, $s$, and $2s$) from layer $l-1$, denoted as $\mathbf{Y}_{s/2}^{l-1}$,  $\mathbf{Y}_{s}^{l-1}$, and $\mathbf{Y}_{2s}^{l-1}$, respectively. Thus, the input $\mathbf{X}_{s}^{l}$ of the $l$-th layer can be formulated as $\mathbf{X}_{s}^{l} = \mathbf{Y}_{s/2}^{l-1} + \mathbf{Y}_{s}^{l} + \mathbf{Y}_{2s}^{l-1}$. Then the aggregated input will be utilized for feature transformation inside the {\em Cell} and {\em Gate}.
%-------------------------------------------------------------------------
\subsubsection{Cell Operation}
With the input $\mathbf{X}_{s}^{l}\in \mathbb{R}^{B\times C\times W\times H}$, we adopt widely-used stacks of separate convolutions as well as identity mapping~\cite{zoph2018learning, liu2018darts, liu2019auto} in each cell without bells-and-whistles. In particular, the hidden state $\mathbf{H}_{s}^{l}\in \mathbb{R}^{B\times C\times W\times H}$ can be represented as
%-------------------------------------------------------------------------
\begin{equation}\label{equ:cell_inside}
\mathbf{H}_s^l=\sum_{O^i\in \mathcal{O}}O^i(\mathbf{X}_s^l)
\end{equation}
%-------------------------------------------------------------------------
where $\mathcal{O}$ indicates the operation set, including SepConv3$\times$3 and identity mapping. Here, operations inside each cell are adopted for fundamental feature aggregation. Then the generated feature map $\mathbf{H}_s^l$ will be transformed to different scales according to the activating factor $\alpha^l_s$. This process will be elaborated in the following section. Moreover, different cell components are compared in Sec.~\ref{sec:cell_component}.
%-------------------------------------------------------------------------
\subsubsection{Soft Conditional Gate}\label{sec:conditional_gate}
The {\em routing probability} of each path is generated from the {\em Gate} function, as presented in the right diagram of Fig.~\ref{fig:arch}. In more detail, we adopt light-weighted convolutional operations in the gate to learn the {\em data-dependent} vector $\mathbf{G}_{s}^{l}$.
%-------------------------------------------------------------------------
\begin{equation}\label{equ:gate_inside}
\mathbf{G}_s^l=\mathcal{F}(\omega^l_{s,2},\mathcal{G}(\sigma(\mathcal{N}(\mathcal{F}(w^l_{s,1},\mathbf{X}_s^l)))))+\beta^l_s
\end{equation}
%-------------------------------------------------------------------------
where $\mathcal{F}(\cdot, \cdot)$ denotes a convolutional function, $\sigma$ indicates $\mathrm{ReLU}$ activation, $\mathcal{N}$ and $\mathcal{G}$ represent batch normalization and global average pooling respectively. Both $\omega$ and $\beta$ are convolutional parameters. Different from traditional RL-based methods~\cite{wang2018skipnet, wu2018blockdrop, veniat2018learning}, which adopt policy gradient to update the agent for discrete path selection, we propose the {\em soft conditional gate} for {\em differentiable} routing. To this end, with the feature vector $\mathbf{G}_{s}^{l}\in \mathbb{R}^{B\times 3\times 1\times 1}$, an activation function~$\delta$ is designed as
%-------------------------------------------------------------------------
\begin{equation}\label{equ:activation_function} 
\delta (\cdot) = \mathrm{max}(0, \mathrm{Tanh}(\cdot))
\end{equation}
%-------------------------------------------------------------------------
Therefore, the activating factor $\alpha^l_s\in \mathbb{R}^{B\times 3\times 1\times 1}$ can be calculated by $\delta(\mathbf{G}_{s}^{l})$, where $\alpha^l_s$ belongs to $[0, 1)$. When $\alpha^l_{s\rightarrow j}=0$, the routing path from scale $s$ to $j$ will be marked as {\em closed}. And all of the paths with $\alpha^l_{s\rightarrow j}>0$ will be reserved, enabling multi-path propagation. To be more specific, the $b$-th input in batch $B$ would generate corresponding  $\alpha^l_{b, s\rightarrow j}\in \mathbb{R}^{1\times 1\times 1\times 1}$, which means the routing paths varies with inputs, or so called {\em data-dependent}. In this way, each path can be taken into consideration individually, rather than only choose the relative important one for propagation~\cite{liu2018darts, wu2019fbnet, liu2019auto}. Furthermore, different activation functions are investigated in Sec~\ref{sec:gating_func}.

With the proposed activation function $\delta$, the transform from scale $s$ to $j\in \{s/2, s, 2s\}$ in the {\em training} process can be formulated as
%-------------------------------------------------------------------------
\begin{equation}\label{equ:transform_train}
\mathbf{Y}_j^l=\alpha^l_{s\rightarrow j}\mathcal{T}_{s\rightarrow j}(\mathbf{H}_s^l)
\end{equation}
%-------------------------------------------------------------------------
where $\mathcal{T}_{s\rightarrow j}$ denotes the scale transformation (including up-sampling, keeping resolution, and down-sampling) from scale $s$ to $j$. Therefore, with the activating factor $\alpha^l_s$, the parameters in $\mathbf{G}_s^l$ will be optimized during back-propagation as long as one path is preserved (namely, $\sum_j\alpha^l_{s\rightarrow j}>0$). 

In the {\em inference} stage, if all of the paths are marked as {\em closed}, the operations in {\em Cell} will be dropped to save computational footprints. Recall from Eq.~\ref{equ:cell_inside}, this process is summarized as
%-------------------------------------------------------------------------
\begin{equation}
\mathbf{H}_s^l=\left\{\begin{array}{lc}\mathbf{X}_s^l&\sum_j\alpha^l_{s\rightarrow j}=0\\\sum_{O^i\in  \mathcal{O}}O^i(\mathbf{X}_s^l)&\sum_j\alpha^l_{s\rightarrow j}>0\end{array}\right.
\end{equation}
%-------------------------------------------------------------------------
\begin{equation}
\mathbf{Y}_j^l=\left\{\begin{array}{lc}0&{\textstyle\sum_j}\alpha^l_{s\rightarrow j}=0,\;j\neq s\\\mathbf{H}_s^l&\textstyle\sum_j\alpha^l_{s\rightarrow j}=0,\;j=s\\\textstyle\alpha^l_{s\rightarrow j}\mathcal{T}_{s\rightarrow j}(\mathbf{H}_s^l)&\textstyle\sum_j\alpha^l_{s\rightarrow j}>0\end{array}\right.
\end{equation}
%-------------------------------------------------------------------------
\subsection{Budget Constraint}\label{sec:budget_constraint}
Considering limited computational resources in the real-world scene, we take the budget constraint into consideration for efficient dynamic routing. Let us denote $\mathcal{C}$ as the computational cost associated to the predefined operation, {\em e.g.,} FLOPs. Recall from Eq.~\ref{equ:cell_inside},~\ref{equ:gate_inside}, and~\ref{equ:transform_train}, we formulate the expected cost inside the node in $s$-th scale and $l$-th layer as
%-------------------------------------------------------------------------
\begin{equation}
\begin{array}{ccl}\mathcal{C}(\mathrm{Node}_s^l)&=&\mathcal{C}(\mathrm{Cell}_s^l)+\mathcal{C}(\mathrm{Gate}_s^l)+\mathcal{C}(\mathrm{Trans}_s^l)\\&=&\mathrm{max}(\alpha_s^l)\sum_{O^i\in \mathcal{O}}\mathcal{C}(O^i)+\mathcal{C}(\mathrm{Gate}_s^l)\\&+&\sum_j\alpha_{s\rightarrow j}^l\mathcal{C}(\mathcal{T}_{s\rightarrow j})\end{array}
\end{equation}
%-------------------------------------------------------------------------
where $\mathrm{Cell}_s^l$, $\mathrm{Gate}_s^l$, and $\mathrm{Trans}_s^l$ indicates the functional operation inside {\em Cell}, {\em Gate}, and {\em Scale Transform}, respectively. Going one step further, the expected cost of the whole routing space could be calculated by
%-------------------------------------------------------------------------
\begin{equation}
\mathcal{C}(\mathrm{Space})=\sum_{l\leq L}\sum_{s\leq1/4}\mathcal{C}(\mathrm{Node}_s^l)
\end{equation}
%-------------------------------------------------------------------------
Then we formulate the expected resource cost $\mathcal{C}(\mathrm{Space})$ into loss function $\mathcal{L}_\mathrm{C}$ for the end-to-end optimization:
%-------------------------------------------------------------------------
\begin{equation}\label{equ:budgeted_constrain}
\mathcal{L}_\mathrm{C}=(\mathcal{C}{(\mathrm{Space})/\mathrm{C}-\mu)}^2
\end{equation}
%-------------------------------------------------------------------------
where $\mathrm{C}$ represents the real resource cost of the whole routing space, and $\mu \in [0, 1]$ indicates the designed attenuation factor. With different $\mu$, the selected routes in each propagation would be adaptively restricted to corresponding budgets. The network performance under different budget constraints will be discussed in Sec.~\ref{sec:resource_budget}.
%-------------------------------------------------------------------------

Overall, the network weights, as well as the soft conditional gates, can be optimized with a joint loss function $\mathcal{L}$ in a unified framework.
%-------------------------------------------------------------------------
\begin{equation}\label{equ:final_loss}
\mathcal{L}=\lambda_1\mathcal{L}_\mathrm{N}+\lambda_2\mathcal{L}_\mathrm{C}
\end{equation}
%-------------------------------------------------------------------------
where $\mathcal{L}_\mathrm{N}$ and $\mathcal{L}_\mathrm{C}$ denotes loss function of the whole network and resource cost, respectively. $\lambda_1$ and $\lambda_2$ are utilized to balance the optimization process of network prediction and resource cost expectation, respectively.
%-------------------------------------------------------------------------
\subsection{Architecture Details}\label{sec:network_arch}
From a macro perspective, we set the depth of routing space to 16 or 33 which is identical with that in widely-used ResNet-50 and ResNet-101~\cite{he2016deep}, namely the total layer $L=16$ or $33$ in Fig.~\ref{fig:arch}. This setting brings convenience to compare with the ResNet-based networks, which could be formulated using the proposed routing space directly. 

When it comes to micro nodes in the network, we adopt three SepConv$3\times3$ in the `STEM' block, where the number of filters is 64 for all of the convolutions. A stride 2 Conv$1\times 1$ is used for all of the $s\rightarrow s/2$ paths, both to reduce feature resolution and double the number of filters. And Conv$1\times 1$ followed by bilinear up-sampling is adopted for all of the $s\rightarrow 2s$ connections, both to increase spatial resolution as well as halve the number of filters. 
% The operations inside each node in well elaborated in Sec.~\ref{sec:routing_process}.

Moreover, a {\em naive decoder} is designed to fuse features for final predictions, which is represented as gray nodes at the end of the network in Fig.~\ref{fig:arch}. Specifically, a Conv$1\times 1$ combined with bilinear up-sampling is used to fuse features from different scales in the decoder. And the prediction in the scale $1/4$ is up-sampled by 4 to generate the final result. The weights in convolutions are initialized with normal distribution~\cite{he2015delving} while the bias $\beta^l_s$ in Eq.~\ref{equ:gate_inside} is initialized to a constant value 1.5 experimentally. When given a budget constraint, we down-sample the input $\mathbf{X}_s^l$ in Eq.~\ref{equ:gate_inside} by 4 times to reduce resource consumption of the gating function. Otherwise, the resolution of input $\mathbf{X}_s^l$ is kept unchanged.
%-------------------------------------------------------------------------
\begin{table*}[t!]
\centering
 \caption{Comparisons with classic architectures on the Cityscapes {\em val} set. `Dynamic' denotes the proposed {\em dynamic routing}. `A', `B', and `C' represent different computational budgets in Sec.~\ref{sec:resource_budget}. `Common' indicates the common connection pattern of corresponding dynamic network. FLOPs$_{Avg}$, FLOPs$_{Max}$, and FLOPs$_{Min}$ represent the {\em Average}, {\em Maximum}, and {\em Minimum} FLOPs of the network, respectively. All of the architectures are sampled from the designed routing space and evaluated by ourself under the same setting.}
\resizebox{\textwidth}{29mm}{
\begin{tabular}{lccccccc}
  \toprule
  Method & Dynamic  & Modeled from  &  mIoU(\%) & FLOPs$_{Avg}$(G) & FLOPs$_{Max}$(G) & FLOPs$_{Min}$(G) & Params(M)\\
  \midrule
  \multirow{4}*{Handcrafted} & \xmark  & FCN-32s~\cite{long2015fully} &  66.9 & 35.1 & 35.1 & 35.1 & 2.9 \\
  						  ~  & \xmark  & DeepLabV3~\cite{chen2017rethinking} & 67.0 & 42.5 & 42.5 & 42.5 & 3.7 \\
  						  ~  & \xmark  & U-Net~\cite{ronneberger2015u} & 71.6 & 53.9 & 53.9 & 53.9 & 6.1 \\
  						  ~  & \xmark  & HRNetV2~\cite{sun2019high} & 72.5 & 62.5 & 62.5 & 62.5 & 5.4 \\

  \midrule
  Searched & \xmark & Auto-DeepLab~\cite{liu2019auto} & 67.2 & 33.1 & 33.1 & 33.1 & 2.5 \\
  \midrule
  Common-A & \xmark & Dynamic-A & 71.6 & 41.6 & 41.6 & 41.6 & 4.1 \\
  Common-B & \xmark & Dynamic-B & 73.0 & 53.7 & 53.7 & 53.7 & 4.3 \\
  Common-C & \xmark & Dynamic-C & 73.2 & 57.1 & 57.1 & 57.1 & 4.5 \\
  \midrule
  {\bf Dynamic-A} & \cmark & Routing-Space & 72.8 & 44.9 & 48.2 & 43.5 & 17.8 \\
  {\bf Dynamic-B} & \cmark & Routing-Space & 73.8 & 58.7 & 63.5 & 56.8 & 17.8 \\
  {\bf Dynamic-C} & \cmark & Routing-Space & 74.6 & 66.6 & 71.6 & 64.3 & 17.8 \\
%  {\bf Dynamic-Raw} & \cmark & Routing-Space & 76.08 & 119.49 & 17.79 \\
  \bottomrule
 \end{tabular}
 \label{tab:dynamic_comparison}
}
\end{table*}
%-------------------------------------------------------------------------
%-------------------------------------------------------------------------
\section{Experiments}\label{sec:experiments}
In this section, we first introduce the datasets and implementation details of the proposed dynamic routing. Then we conduct abundant ablation studies on the Cityscapes dataset~\cite{cordts2016cityscapes}. And detailed analyses will be given to reveal the effect of each component. Finally, comparisons with several benchmarks on the Cityscapes~\cite{cordts2016cityscapes} and PASCAL VOC 2012 dataset~\cite{everingham2010pascal} will be reported to illustrate the effectiveness and the efficiency of the proposed method.

%-------------------------------------------------------------------------
\subsection{Datasets}
\noindent{\bf Cityscapes:} The Cityscapes~\cite{cordts2016cityscapes} is a widely used dataset for urban scene understanding, which contains 19 classes for evaluation. The dataset involves 5000 {\em fine} annotations with size $1024\times 2048$, which can be divided into 2975, 500, and 1525 images for training, validation, and testing, respectively. It has another 20k {\em coarse} annotations for training, which are {\em not used} in our experiments.

\noindent{\bf PASCAL VOC:} We carry out experiments on the PASCAL VOC 2012 dataset~\cite{everingham2010pascal} that includes 20 object categories and one background class. The original dataset contains 1464, 1449, and 1456 images for training, validation, and testing, respectively. Here, we use the augmented data provided by~\cite{hariharan2011semantic}, resulting in 10582 images for training.
%-------------------------------------------------------------------------
\subsection{Implementation Details}
Herein, optimization details are reported for convenient implementation. For better performance, the factor $\lambda_1$ in Eq.~\ref{equ:final_loss} is set to 1.0. And $\lambda_2$ is set according to different budget constraints in Sec.~\ref{sec:resource_budget}. The network optimization is conducted using SGD with weight decay $1e^{-4}$ and momentum 0.9. Similar to~\cite{chen2017rethinking, yu2018bisenet, song2019learnable}, we adopt the `poly' schedule where the initial rate is multiplied by $(1-\frac{iter}{iter_{max}})^{power}$ in each iteration with power 0.9. In training stage, we randomly flip and scale each image by 0.5 to 2.0$\times$. Different initial rates are applied according to the experimental setting. Specifically, we set initial rate to 0.05 and 0.02 when training from scratch and using ImageNet~\cite{deng2009imagenet} pre-training, respectively. For Cityscapes~\cite{cordts2016cityscapes}, we construct each minibatch for training from 8 random $768\times 768$ image crops. For PASCAL VOC 2012~\cite{everingham2010pascal}, 16 random $512\times 512$ image crops are adopted for optimization in each iteration. 

%------------------------------------------------------------------------
\begin{figure}[t!]
  \centering
    \subfigure[Network architecture of Common-A]{
    \includegraphics[width=0.98\linewidth]{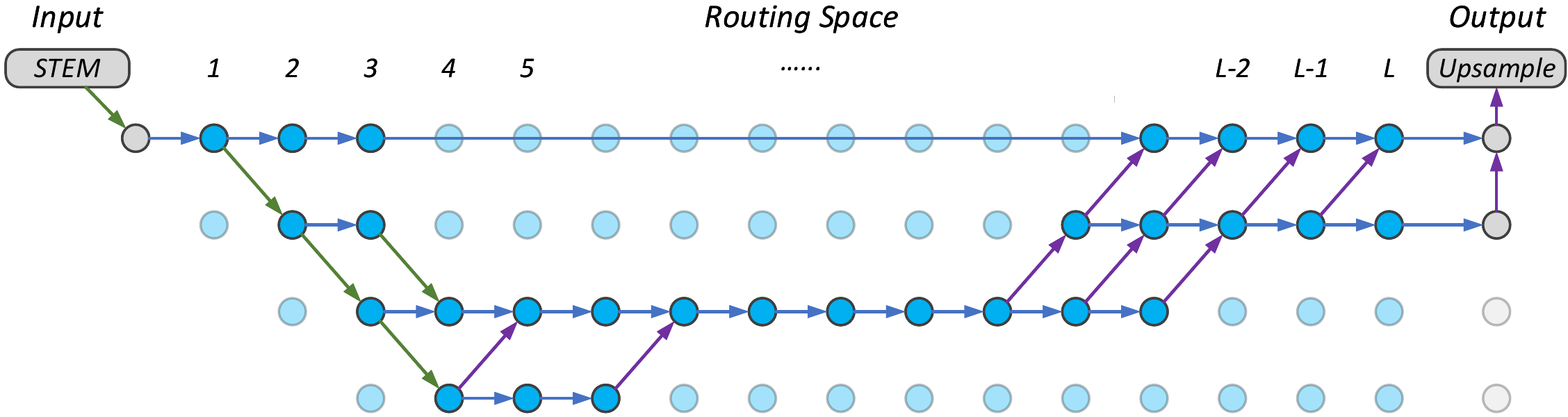}
    \label{fig:archs_in_space_common_a}
    }
    \subfigure[Network architecture of Common-B]{
    \includegraphics[width=0.98\linewidth]{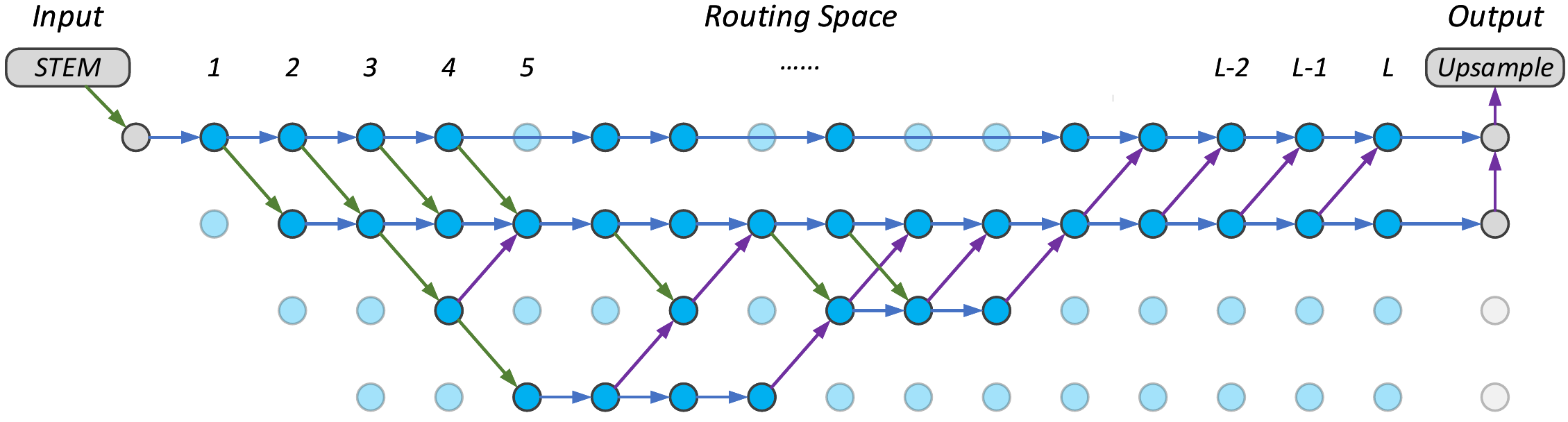}
    \label{fig:archs_in_space_common_b}
    }
    \subfigure[Network architecture of Common-C]{
    \includegraphics[width=0.98\linewidth]{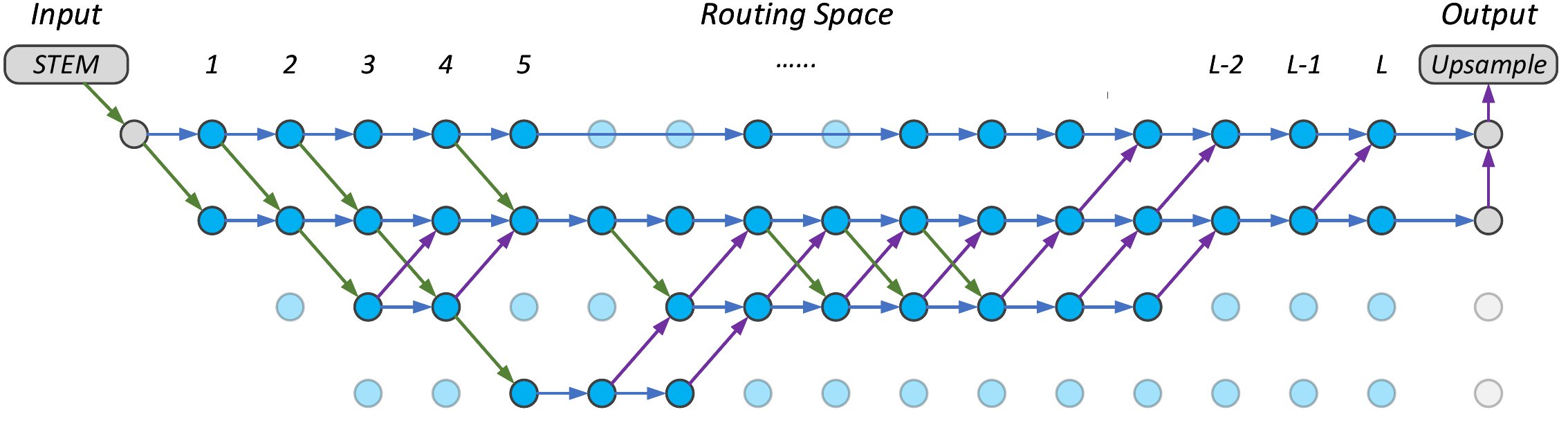}
    \label{fig:archs_in_space_common_c}
    }
\caption{Network architectures of Common-A, B, C, which are extracted from Dynamic models with different budget constraints in Tab.~\ref{tab:dynamic_comparison}, are visualized in~\ref{fig:archs_in_space_common_a},~\ref{fig:archs_in_space_common_b}, and~\ref{fig:archs_in_space_common_c}, respectively.}
\label{fig:arch_in_space_common}
\end{figure}
%------------------------------------------------------------------------
\subsection{Dynamic Routing}\label{sec:exper_dynamic_routing}
%-------------------------------------------------------------------------
To demonstrate the superiority of the dynamic routing, we compare the dynamic networks with several existing architectures and static routes sampled from the routing space. In particular, traditional {\em human-designed networks} as well as {\em searched architectures}, including FCN-32s~\cite{long2015fully}, U-Net~\cite{ronneberger2015u}, DeepLabV3~\cite{chen2017rethinking}, HRNetV2~\cite{sun2019high}, and Auto-DeepLab~\cite{liu2019auto}, are modeled in the routing space with similar connection patterns, as visualized in Fig.~\ref{fig:arch_in_space}. For fair comparisons, we align the computational overhead with these methods by giving different budget constraints to the loss function in Eq.~\ref{equ:budgeted_constrain}. Consequently, three types of dynamic networks can be generated (please refer to Sec.~\ref{sec:resource_budget} for details), denoted as Dynamic-A, B, and C in Tab.~\ref{tab:dynamic_comparison}. Compared with the handcrafted and searched architectures, the proposed dynamic routing achieves much better performance under similar costs. For instance, given the budget constraint around 45G, 55G, and 65G, the Dynamic-A, B, and C attain {\bf 5.8\%},  {\bf 2.2\%}, and {\bf 2.1\%} absolute gain over the modeled DeepLabV3, U-Net, and HRNetV2, respectively.

Furthermore, the fundamental routes of dynamic networks, which are preserved over 95\% forward inference, are extracted to formulate corresponding {\em common network}. The connection patterns of common networks are presented in Fig.~\ref{fig:arch_in_space_common}. We further compare the dynamic networks with common architectures (Common-A, B, and C) in Tab.~\ref{tab:dynamic_comparison}. Specifically, with the dynamic routing framework, the {\em dynamic} network will have better performance than the static {\em common} one under each budget constraint. This can be concluded from the comparisons in Tab.~\ref{tab:dynamic_comparison}. 

We observe that the connected routes of common networks in Fig.~\ref{fig:arch_in_space_common} share a {\bf similar tendency} with several known architectures, {\em e.g.,} the human-designed U-Net~\cite{ronneberger2015u} and the NAS-based Auto-DeepLab~\cite{liu2019auto}. In particular, down-sampling operations are adopted in the front part, and up-sampling operations are preferred in the latter part of the network. Moreover, high-resolution features in the low-level stage are needed for object details (visualized in Fig.~\ref{fig:arch_intro}), which may lead to better performance.
%-------------------------------------------------------------------------
\subsection{Component-wise Analysis}
To reveal the effect of each component in the proposed method, we will decompose our approach step-by-step in this section. Firstly, the {\em components} inside cells will be discussed in detail. Then we investigate the {\em activation function} of the proposed soft conditional gate. The effect of different {\em resource budgets} will be further illustrated in the end.
%-------------------------------------------------------------------------
\subsubsection{Cell Component}\label{sec:cell_component}
For fair comparisons with previous architectures, only the basic convolutional operations and identity mapping are used inside each cell without bells-and-whistles. Experimental results with several classic operations including BottleNeck~\cite{he2016deep}, MBConv~\cite{sandler2018mobilenetv2}, and SepConv~\cite{chollet2017xception} are presented in Tab.~\ref{tab:cell_comparison}. We find the dynamic network achieve the best performance when stacking two SepConv3$\times$3 for feature transform and heavier operations contribute no more gain. We guess this could be attributed to the reason that the routing architecture plays a more significant role than heavier operations. Indeed, we also conduct experiments with larger kernels ({\em e.g.,} SepConv5$\times$5) when the resolution scale is 1/4, but only find 0.2\% absolute gain. Thus, we only use SepConv3$\times$3 in this paper for simplicity.
%------------------------------------------------------------------------
\begin{table}[t!]
\centering
 \caption{Comparisons among different cell components on the Cityscapes {\em val} set. `$\times 2$' and `$\times 3$' mean stacking 2 and 3 SepConv3$\times$3, respectively. Due to the data-dependent property of the dynamic routing, we report the {\em average} FLOPs here.}
\resizebox{0.45\textwidth}{14mm}{
\begin{tabular}{lccc}
  \toprule
  Cell Operation  &  mIoU(\%) & FLOPs(G) & Params(M)\\
  \midrule
  BottleNeck~\cite{he2016deep} & 73.7 & 1134.8 & 203.9 \\
  MBConv~\cite{sandler2018mobilenetv2} & 75.0 & 323.8 & 48.2 \\
  SepConv3$\times$3  & 71.2 & 81.4 & 12.6 \\
  SepConv3$\times$3 $\times 2$ & 76.1 & 119.5 & 17.8 \\
  SepConv3$\times$3 $\times 3$ & 75.2 & 153.8 & 22.9 \\
  \bottomrule
 \end{tabular}
 }
 \label{tab:cell_comparison}
\end{table}
%-------------------------------------------------------------------------
\begin{table}[t!]
\centering
 \caption{Comparisons among different activation functions on the Cityscapes {\em val} set. Due to the data-dependent property of the dynamic routing, we report the {\em average} FLOPs here.}
\resizebox{0.45\textwidth}{14mm}{
\begin{tabular}{lccc}
  \toprule
  Activation  &  mIoU(\%) & FLOPs(G) & Params(M)\\
  \midrule
  Fix & 74.5 & 103.1 & 15.3 \\
  \midrule
  Softmax & 74.1 & 120.0 & 17.8 \\
  Sigmoid & 75.9 & 120.0 & 17.8 \\
  max(0, Tanh) & 76.1 & 119.5 & 17.8 \\
  \bottomrule
 \end{tabular}
 }
 \label{tab:activation_compare}
\end{table}
%---------------------------------------------------------------------
%------------------------------------------------------------------------
 \begin{table}[t!]
 \centering
  \caption{Comparisons among different resource budgets on the Cityscapes {\em val} set. $\lambda_2$ and $\mu$ denote the coefficients for budget constraint in Sec.~\ref{sec:budget_constraint}. Due to the data-dependent property of the dynamic routing, we report the {\em average} FLOPs here.}
 \resizebox{0.48\textwidth}{14mm}{
 \begin{tabular}{lcccc}
   \toprule
   Method & $\lambda_2/\mu$ & mIoU(\%) & FLOPs(G) & Params(M)\\
   \midrule
   Network-Fix  & - & 74.5 & 103.1 & 15.3 \\
   \midrule
   Dynamic-A & 0.8/0.1 & 72.8 & 44.9 & 17.8 \\
   Dynamic-B & 0.5/0.1 & 73.8 & 58.7 & 17.8 \\
   Dynamic-C & 0.5/0.2 & 74.6 & 66.6 & 17.8 \\
   Dynamic-Raw & 0.0/0.0 & 76.1 & 119.5 & 17.8 \\
   \bottomrule
  \end{tabular}
 }
  \label{tab:resource_budget}
 \end{table}
%-------------------------------------------------------------------------
\subsubsection{Activation Function}\label{sec:gating_func}
We further compare several widely-used activation functions of the proposed soft conditional gate in Sec.~\ref{sec:conditional_gate}. Firstly, all of the paths in the routing space are fixed with no difference to formulate our baseline, namely the `Fix' in Tab.~\ref{tab:activation_compare}. Then, the activation function $\delta$ in Eq.~\ref{equ:activation_function} is replaced by the candidate in Tab.~\ref{tab:activation_compare} directly. We find the proposed $\mathrm{max}(0, \mathrm{Tanh})$ achieves better performance than others. What's more, the performance of the $\mathrm{Softmax}$ activation, which considers three routing paths in each cell together, is inferior to that of considered individually, {\em e.g.,} $\mathrm{Sigmoid}$ and $\mathrm{max}(0, \mathrm{Tanh})$. This means each path should be decoupled in the soft conditional gate. Then, the paths with activating factor $\alpha>0$ would be preserved during this forward inference, as elaborated in Sec.~\ref{sec:conditional_gate}.
%-------------------------------------------------------------------------
\begin{figure}[t!]
  \centering
    \caption{Distribution of route activating probabilities in dynamic networks. Most of the paths tend to be preserved without budget constraints in Dynamic-Raw. Given resource budgets, different proportions of routes will be {\em closed} in Dynamic-A, B, and C.}
  \includegraphics[width=0.95\linewidth]{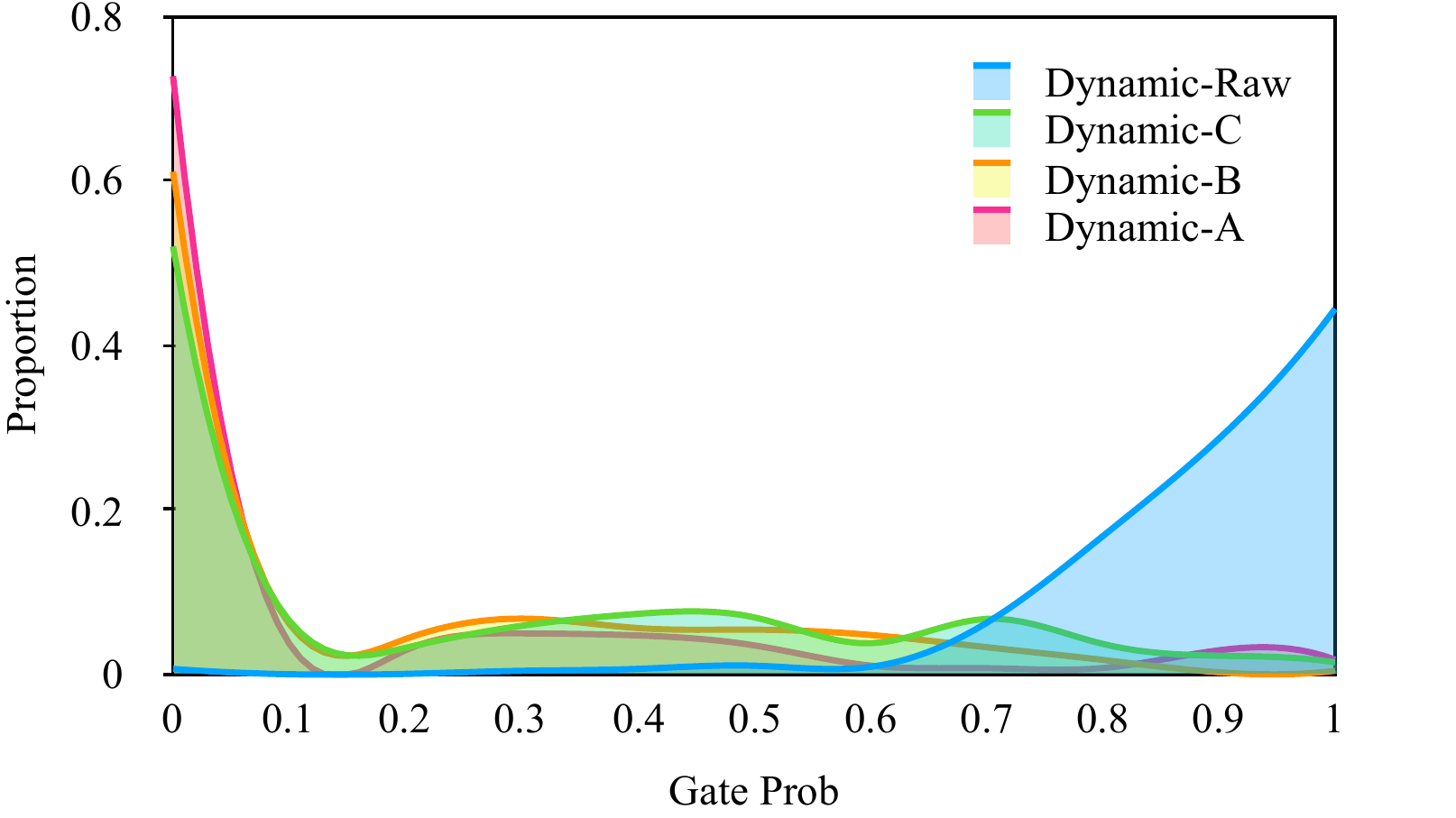}\\

  \label{fig:gate_distribution}
\end{figure}
%-------------------------------------------------------------------------
\begin{table}[t!]
\centering
 \caption{Experiments with different settings on Cityscapes {\em val} set with a single scale and no flipping. `ImageNet' denotes ImageNet pre-training. `SDP' indicates Scheduled Drop Path.}
\resizebox{0.45\textwidth}{22.5mm}{
\begin{tabular}{lcccc}
  \toprule
  Method  & Iter(K) & ImageNet & SDP &  mIoU(\%) \\
  \midrule
  Network-Fix & 186 & \xmark & \xmark & 74.5 \\
  {\bf Dynamic} & 186 & \xmark & \xmark & 76.1 \\
  \midrule
  Network-Fix & 372 & \xmark & \xmark & 76.3 \\
  {\bf Dynamic} & 372 & \xmark & \xmark & 77.4 \\
  \midrule
  Network-Fix & 558 & \xmark & \cmark & 76.7 \\
  {\bf Dynamic} & 558 & \xmark & \cmark & 78.3 \\
  \midrule
  Network-Fix & 186 & \cmark & \xmark & 75.8 \\
  {\bf Dynamic} & 186 & \cmark & \xmark & {\bf 78.6} \\
  \bottomrule
 \end{tabular}
 \label{tab:result_city_iter}
}
\end{table}
%------------------------------------------------------------------------
%-------------------------------------------------------------------------
\begin{table*}[th!]
\centering
 \caption{Comparisons with previous works on the Cityscapes. mIoU$_{test}$ and mIoU$_{val}$ denote performance on {\em test} set and {\em val} set respectively. Multi-scale and flipping strategy are used in {\em test} set but dropped in {\em val} set. We report FLOPs with input size $1024\times 2048$. }
\begin{threeparttable}
\begin{tabular}{lccccc}
  \toprule
  Method  & Backbone & mIoU$_{test}$(\%) & mIoU$_{val}$(\%) & FLOPs(G) \\ % & Params(M) \\
  \midrule
  BiSenet~\cite{yu2018bisenet} & ResNet-18 & 77.7 & 74.8 & 98.3$^\dagger$ \\
  DeepLabV3~\cite{chen2017rethinking} & ResNet-101-ASPP & - & 78.5 & 1778.7 \\ % & & 58.00 \\
  Semantic FPN~\cite{kirillov2019panoptic} & ResNet-101-FPN & - & 77.7 & 500.0 \\
%  Semantic FPN~\cite{kirillov2019panoptic} & ResNeXt-101-FPN & - & 79.1 & 800.0 \\
  DeepLabV3+~\cite{chen2018encoder} & Xception-71-ASPP & - & 79.6 & 1551.1 \\ % & & 58.00 \\
  PSPNet~\cite{zhao2017pyramid} & ResNet-101-PSP & 78.4 & 79.7 & 2017.6  \\ % & & 65.90 \\
  Auto-DeepLab*~\cite{liu2019auto} & Searched-F20-ASPP & 79.9 & 79.7 & 333.3 \\ % & & 10.15 \\
  Auto-DeepLab*~\cite{liu2019auto} & Searched-F48-ASPP & 80.4 & 80.3 & 695.0 \\ % & & 10.15 \\
  \midrule
  {\bf Dynamic}* & Layer16 & 79.1 & 78.3 & 111.7 \\
  {\bf Dynamic} & Layer16 & 79.7 & 78.6 & 119.4 \\ % 80.3 & 78.9\\
  {\bf Dynamic} & Layer33 & 80.0 & 79.2 & 242.3 \\ % & & 38.87\\
  {\bf Dynamic} & Layer33-PSP & {\bf 80.7} & {\bf 79.7} & {\bf 270.0} \\ % & & 39.07\\
  \bottomrule
 \end{tabular}
\begin{tablenotes}
        \footnotesize
        \item[$\dagger$] estimated from corresponding settings\
        \item[*] training from scratch
\end{tablenotes}
\end{threeparttable}
\label{tab:result_city_val_test}
\end{table*}
%------------------------------------------------------------------------
\begin{table*}[th!]
\centering
 \caption{Comparisons with previous works on the PASCAL VOC 2012. mIoU$_{test}$ and mIoU$_{val}$ denote performance on {\em test} set and {\em val} set respectively. Multi-scale and flipping strategy are used in {\em test} set but dropped in {\em val} set. We report FLOPs with input size $512\times 512$.}
\begin{threeparttable}
\begin{tabular}{lcccc}
  \toprule
  Method  & Backbone & mIoU$_{test}$(\%) & mIoU$_{val}$(\%) & FLOPs(G) \\ % & Params(M) \\
  \midrule
  DeepLabV3~\cite{chen2017rethinking} & MobileNet-ASPP & - & 75.3 & 14.3 \\ % & 11.15 \\
  DeepLabV3~\cite{chen2017rethinking} & MobileNetV2-ASPP & - & 75.7 & 5.8 \\ % & 4.52 \\
  Auto-DeepLab~\cite{liu2019auto} & Searched-F20-ASPP & 82.5 & 78.3 & 41.7$^\dagger$ \\ % & 10.15 \\
  \midrule
%   Network-Fix & Layer16 & 77.4 & 12.9 \\
  {\bf Dynamic} & Layer16 & 82.8 & 78.6 & 14.9 \\ % & 17.79\\
  {\bf Dynamic} & Layer33 & {\bf 84.0} & {\bf 79.0} & {\bf 30.8} \\ % & -\\
  \bottomrule
\end{tabular}
\begin{tablenotes}
        \footnotesize
        \item[$\dagger$] estimated from corresponding settings
\end{tablenotes}
\end{threeparttable}
\label{tab:result_voc_val_test}
\end{table*}
%------------------------------------------------------------------------

\subsubsection{Resource Budgets}\label{sec:resource_budget}
With the designed gating function, we give different resource budgets by adjusting the coefficient $\lambda_2$ and $\mu$. As presented in Tab.~\ref{tab:resource_budget}, the routing framework will generate several types of dynamic networks (Dynamic-A, B, and C) if given different budget constraints. Compared with the raw dynamic network without resource budget (Dynamic-Raw), the cost in Dynamic-C is reduced to {\bf 55.7\%} with little performance drop. Meanwhile, the Dynamic-C still outperforms the fully-connected Network-Fix both on effectiveness and efficiency. And the resource cost can be further reduced to {\bf 37.6\%} (Dynamic-A) with stronger constraints.

Moreover, we present the distribution of route activating probabilities in Fig.~\ref{fig:gate_distribution}. It is clear that most of the paths tend to be preserved in Dynamic-Raw. Different proportions of routes will be dropped if given resource budgets, which can be learned from distributions in Fig~\ref{fig:gate_distribution}. Consequently, the dynamic routing would cut out useless paths as well as cells during inference. We find the gap between FLOPs$_{Max}$ and FLOPs$_{Min}$ in Tab.~\ref{tab:dynamic_comparison} is relatively small (10\%), which can be attributed to the effect of budget constraint. Indeed, we also tried different types of coefficient of variation~\cite{shazeer2017outrageously} to enlarge the gap, but found inferior performance.

%We give more optimizing details and qualitative visualizations in the {\em supplementary material}.
%------------------------------------------------------------------------
\subsection{Experiments on Cityscapes}
We carry out experiments on the Cityscapes~\cite{cordts2016cityscapes} dataset using {\em fine} annotations only. In Tab.~\ref{tab:result_city_iter}, we compare the dynamic network with the fixed backbone under several training settings on the {\em val} set. The proposed method achieves consistent improvement in different situations. With the Scheduled Drop Path~\cite{zoph2018learning, liu2019auto} and ImageNet~\cite{deng2009imagenet} pre-training, the performance of the dynamic network ($L=16$) can be further improved. Comparisons with several previous works are given in Tab.~\ref{tab:result_city_val_test}. With the {\em similar resource cost}, the proposed dynamic network attains 78.6\% mIoU on the {\em val} set, which achieves a {\bf 3.8\%} absolute gain over the well-designed BiSenet~\cite{yu2018bisenet}. With the simple scale transform modules without bells-and-whistles, the dynamic network ($L=33$) achieves comparable performance with the state-of-the-art but consumes much fewer cost. Moreover, in conjunction with the context capturing module ({\em e.g.,} PSP block), the proposed method has further improvements and achieves {\bf 80.7\%} mIoU on the Cityscapes {\em test} set.
%------------------------------------------------------------------------
\subsection{Experiments on PASCAL VOC}
%---------------------------------------------------------------------
We further compare with similar methods (pre-trained on the COCO~\cite{lin2014microsoft} dataset), which focus on the architecture design with comparable computational overhead, on the PASCAL VOC 2012~\cite{everingham2010pascal} dataset. In particular, the proposed approach surpasses Auto-DeepLab~\cite{liu2019auto}, which would cost 3 GPU days for architecture searching, in both accuracy and efficiency, as shown in Tab.~\ref{tab:result_voc_val_test}. Compared with the MobileNet-based DeepLabV3~\cite{chen2017rethinking}, the dynamic network still attains better performance with a similar resource cost. 
%---------------------------------------------------------------------
\section{Conclusion}
In this work, we present the dynamic routing for semantic segmentation. The key difference from prior works lies in that we generate {\em data-dependent} forward paths according to the scale distribution of each image. To this end, the soft conditional gate is proposed to select scale transformation routes in an end-to-end manner, which will learn to drop useless operations for efficiency if given resource budgets. Extensive ablation studies have been conducted to demonstrate the superiority of the dynamic network over several static architectures, which can be modeled in the designed routing space. Experiments on Cityscapes and PASCAL VOC 2012 prove the effectiveness of the proposed method, which achieves comparable performance with state-of-the-arts but consumes much fewer computational resources.
%
%Our research delivers an important message: the dynamic property can be utilized to generate {\em data-dependent} network architectures for semantic representation, so that each image can be handled in the most appropriate manner. 

\section*{Acknowledgement}
This work was supported by National Key Research and Development Program of China 2018YFD0400902 and National Natural Science Foundation of China 61573349.

\clearpage

{\small
\bibliographystyle{ieee_fullname}
\bibliography{main}
}

\end{document}